\documentclass[default,iicol,sn-basic]{sn-jnl}% Default with double column layout

\usepackage{mathtools}
\usepackage{graphicx,grffile}     % For images, image paths which contain '.' before the extension
\usepackage[capitalise]{cleveref} % For automatically putting Figure/Table/etc before references
\usepackage{amsmath}
\usepackage{booktabs}
\usepackage{tabularx}
\usepackage{icomma}        % To remove space after comma in math mode.
\usepackage{mathptmx}      % use Times fonts if available on your TeX system
\makeatletter
\newcases{lrdcases}
{\quad}
{$\m@th\displaystyle{##}$\hfil}
{$\m@th\displaystyle{##}$\hfil}
{\lbrace}
{} % {\rbrace}

\newcases{lrdcases*}
{\quad}
{$\m@th\displaystyle{##}$\hfil}
{{##}\hfil}
{\lbrace}
{} %{\rbrace}
\makeatother
\newcommand{\IGNORE}[1]{}

\newcommand{\TheSwarm}{S}
\newcommand{\TheSwarmSize}{N}
\newcommand{\SwarmDensity}{\rho_{S}}

\newcommand{\SwarmNSearching}{\TheSwarmSize_{s}(t)}
\newcommand{\SwarmNHoming}{\TheSwarmSize_{h}(t)}
\newcommand{\SwarmNAvoiding}{\TheSwarmSize_{av}(t)}
\newcommand{\SwarmNAvoidingOneRobot}{\TheSwarmSize_{av}^1(t)}

\newcommand{\SwarmNAvoidingWhileHoming}{\TheSwarmSize^h_{av}(t)}
\newcommand{\SwarmNAvoidingWhileSearching}{\TheSwarmSize^s_{av}(t)}
\newcommand{\NBlocksInEnv}[1]{B(#1)}
\newcommand{\CRWDiffusionD}[1]{F(#1)}
\newcommand{\TheScenario}{m}
\newcommand{\AllScenarios}{M}
\newcommand{\ScenarioHeterogeneity}[1]{C_{df}(#1)}
\newcommand{\ScenarioCACharacterization}[1]{C_{ca}(#1)}
\newcommand{\DiffusionTheta}{{D}_{\theta}}
\newcommand{\DiffusionXY}{{D}_{xy}}
\newcommand{\QueueHoming}{Q_h}
\newcommand{\QueueCA}{Q_{ca}}
\newcommand{\EnvHeterogeneity}[1]{H(#1)}
\newcommand{\OperatingArea}{A}
\newcommand{\ArenaXY}{\mathbf{x}}
\newcommand{\NestXY}{\mathbf{x}_n}
\newcommand{\SubAreaCenter}{\mathbf{c}_j}
\newcommand{\SubAreaBlockDensity}{\rho_{j}}
\newcommand{\SubAreaNBlocks}[1]{B_j(#1)}
\newcommand{\ExpectedAcqX}{E[x_{acq}]}

\newcommand{\ExpectedAcqXY}{E[\mathbf{x}_{acq}]}
\newcommand{\dtwonorm}[1]{\left\lVert#1\right\rVert}

\newcommand{\figroot}[1]{figures/#1}

\title[Non-Linear Swarm Behaviors]%
{Characterizing the Limits of Linear Modeling of Non-Linear Swarm Behaviors
}
\author*[1]{\fnm{John} \sur{Harwell}}\email{harwe006@umn.edu}
\author[1]{\fnm{Angel} \sur{Sylvester}}\email{sylve057@umn.edu}
\author[1]{\fnm{Maria} \sur{Gini}}\email{gini@umn.edu}

\affil[1]{\orgdiv{Department of Computer Science \& Engineering},
          \orgname{University of Minnesota},
          \orgaddress{\street {200 Union St SE},
          \city{Minneapolis}, \postcode{55455}, \state{MN}, \country{USA}}}

\date{Received: date / Accepted: date}
\begin{document}

\abstract{We study the limits of linear modeling of behavior of robot swarms by
  characterizing the inflection point beyond which linear models of swarm
  collective behavior break down. The problem we consider is a central place
  object gathering task. We design a linear model which strives to capture the
  underlying dynamics of object gathering from first principles, rather than
  extensively relying on post-hoc model fitting.  We demonstrate in simulation
  that our model accurately captures underlying linear swarm behavioral dynamics
  in two cases: when the swarm can be approximated using the mean-field model
  (large swarms), and when it cannot, and finite-size effects are present (small
  swarms).  We further apply our model to swarms exhibiting non-linear
  behaviors, and show that it still provides accurate predictions in some
  scenarios, thereby establishing better practical limits on linear modeling of
  swarm behaviors.  }


\maketitle
%%%%%%%%%%%%%%%%%%%%%%%%%%%%%%%%%%%%%%%%%%%%%%%%%%%%%%%%%%%%%%%%%%%%%%%%%%%%%%%%
% Introduction
%%%%%%%%%%%%%%%%%%%%%%%%%%%%%%%%%%%%%%%%%%%%%%%%%%%%%%%%%%%%%%%%%%%%%%%%%%%%%%%%
%
\section{Introduction}\label{sec:intro}
Swarm Robotics (SR) is the study of the coordination of large numbers of simple
robots~\citep{Sahin2005}. SR systems can be homogeneous, i.e., single robot type
and identical control software, or heterogeneous, i.e., multiple robot types and
control software~\citep{%%Dorigo2013,Rizk2019,
prorok2017impact,
Ramachandran2020}. The main
differentiating factors between SR systems and multi-agent robotics systems stem
from the mechanisms on which SR systems are based.  Historically, these were
principles of biological mimicry or problem solving techniques inspired by
natural systems of agents such as bees, ants, and termites~\citep{Labella2006},
though modern SR systems typically incorporate more mathematically rigorous
elements of conventional multi-robot system design~\citep{Castello2016}.

% Development of effective strategies for SR systems to solve problems such as
% foraging, collective transport of heavy objects, environmental
% monitoring/cleanup~\citep{Schill2016a}, collective decision making,
% surveillance~\citep{Borko2016a}, and construction~\citep{Petersen2011} can be
% accelerated by drawing inspiration from natural systems, due to their shared
% properties (e.g., scalability, emergent self-organization, flexibility,
% robustness)~\citep{Sahin2005,Hecker2015,Harwell2020a}. Furthermore, SR systems
% with these properties are well suited to tackle tasks in dangerous
% environments when robustness and flexibility are key to success, such as
% space, tracking lake health, clearing space in mining, hazardous material
% cleanup, search and
% rescue~\citep{Rouff2004,Sahin2005,Hecker2015,Labella2006,Kumar2003,Flushing2014}.

% [JRH] Address ijcai#7 comment about not talking about the task before
% presenting the equations.

In this work, we study the \emph{central place foraging problem} in which robots
gather objects (blocks) across a finite operating arena and bring them to a
central location (nest) under various conditions and constraints.  Foraging is
one of the most studied applications of SR, due to its straightforward mapping
to real-world applications~\citep{Hecker2015}; for an extensive discussion of
the state of the art, see~\citep{Lu2020}.  The complexity of the foraging task
frequently gives rise to non-linear behavioral dynamics caused by inter-robot
interactions.

It has been established that above a certain \emph{swarm density}
$\SwarmDensity$ (ratio of swarm size to arena
size)~\citep{Sugawara1997,Hamann2013}, interactions between robots can produce
non-linear behaviors which are not predictable from the individual components,
e.g., from the swarm control
algorithm~\citep{Cotsaftis2009,George2005,Hunt2020,DeWolf2005}. Hence, the
collective performance of a swarm $\TheSwarm$ of $\TheSwarmSize$ robots each
running an identical control algorithm $\kappa$ can be a non-linear function of
the behavior of a system of $\TheSwarmSize$ independent
robots~\citep{Harwell2020a}.  Below this density threshold, swarm behavior can
be well approximated using linear models.

% [JRH] Address ijcai#10,63 comment low density swarms not being swarms.

Many real problems across scales can be tackled using swarms in low density
environments independently of swarm size, in which each robot is responsible for
a large area on the order of $100m^2$; these include indoor warehouses
($\approx{64}m^2$), outdoor search and rescue, precision agriculture, and field
monitoring ($600,000m^2\approx{150}$ acres). Recent results summarizing the
challenges of moving swarms into the real world argue that directing research
towards low density swarm applications is critical~\citep{Tarapore2020}. Some
researchers hold that swarms with low $\SwarmDensity$ are not properly
characterized as swarms, and are instead systems of independent robots because
they lack the high level of inter-agent interaction which characterizes natural
swarms.  We argue that a high level of inter-agent interaction, and therefore
potentially emergent self-organization, is only one of the defining properties
of swarms. Low density swarms exhibiting the properties of scalability,
flexibility, and robustness can properly still be considered
swarms~\citep{Harwell2020a}.
%
% These behaviors can be caused by: (1) excessive inter-robot interference,
% which results in superlinear performance losses~\citep{Harwell2020a}, and (2)
% \emph{emergent self-organization}, which occurs as agents in SR systems work
% collectively to find solutions to problems unsolvable by individual robots,
% and generally results in superlinear performance
% gains~\citep{Winfield2005a,Galstyan2005}.
%
The exact value of $\SwarmDensity$ at which a given linear model of swarm
behavior breaks down is influenced by many factors, including the control
algorithm $\kappa$, the number of robots $\TheSwarmSize$, and characteristics of
the problem being solved, so in general it cannot be determined \emph{a priori}.

In this work we seek to characterize, as a function of $\SwarmDensity$, the
\emph{practical} limits of using linear models to model non-linear behaviors of
a swarm engaged in real-world foraging. Specifically, we will show that
non-linear behavioral dynamics in the foraging problem can be captured using
linear modeling which is derived directly from problem features, such as swarm
density, swarm size, and robot control algorithm.

% [JRH] Cut for space
%% The rest of this paper is organized as
%% follows. In~\cref{sec:motivation,sec:background}, we motivate this work by
%% discussing trends and related work in SR system design, and the differential
%% equation paradigm on which our model is based. In~\cref{sec:method} we develop
%% our model. In~\cref{sec:exp-defs} we define the experiments to evaluate our
%% model and test its limits across multiple spatial object distributions, scales,
%% and densities. Our results in~\cref{sec:results} show that our model provides
%% accurate predictions across block distributions, scales, and densities,
%% including some in which the swarm itself is exhibiting non-linear behaviors,
%% showing its broad utility in future SR system design.

%%%%%%%%%%%%%%%%%%%%%%%%%%%%%%%%%%%%%%%%%%%%%%%%%%%%%%%%%%%%%%%%%%%%%%%%%%%%%%%%
% Related Work
%%%%%%%%%%%%%%%%%%%%%%%%%%%%%%%%%%%%%%%%%%%%%%%%%%%%%%%%%%%%%%%%%%%%%%%%%%%%%%%%
%
\section{Motivation and related work}\label{sec:motivation}
Many SR systems have been designed around imitating natural systems, and use
heuristic decision making~\citep{Castello2016} rather than combining natural
principles with a strong mathematical
grounding~\citep{Talamali2020}. Nevertheless, heuristic approaches to swarm
control have been effective for robots that operate with incomplete information
and limited computing power. SR researchers taking this approach average large
numbers of simulation runs to develop more accurate models of swarm behavior and
obtain empirical insights into real-world problems~\citep{Harwell2019a}.  The
emphasis on empirical rather than mathematical models has been a chief
impediment to a wider use of SR systems~\citep{Lerman2004a}.  Systematically
varying individual agent parameters to study their effect on collective swarm
behavior is impractical, even in simulation. Mathematical characterization of
collective swarm behavior is more difficult, but provides the means to precisely
predict it \emph{a priori}---\emph{without} the need of repetitive experiments.

Given the complexity of SR systems, and the frequently non-linear ways in which
behaviors can arise, it is difficult to obtain precise bounds on the collective
behavior of a swarm $\TheSwarm$ of $\TheSwarmSize$ robots each running a control
algorithm $\kappa$. Robots might need to respond to environmental signals that
arrive at unpredictable times; such systems are well-modeled as asynchronous,
and therefore difficult to predict precisely.  However, if we conceptualize
$\TheSwarm$ as a differentiable, continuous quantity, its dynamics can be
modeled with Ordinary Differential Equations (ODEs) whose variables are the
population counts associated with the different roles. We can apply a
macroscopic-continuous ODE modeling approach for the \emph{average} behavior of
$\TheSwarm$ in the steady state~\citep{Berman2007}, with the caveat that when
using such a model to determine the behavior of $\TheSwarm$, actual system
behaviors could be far from the average~\citep{Lerman2004a}. Usually, the larger
the system, the smaller the fluctuations; in small systems the fluctuations
resulting from \emph{finite size effects} can be of order $\TheSwarmSize$,
resulting in models which are accurate at asymptotically large scales but not at
small scales. The master equation~\citep{VanKampen2007}, which is typically used
to model the expected average behavior of systems, can be used to calculate the
deviation from the average, but such calculations are often intractable or
algebraically difficult.

A promising mathematically rigorous methodology utilizing macro- and microscopic
models for group dynamics and individual behavior over
% [JRH] Thesis:
% ~\citep{Dorigo2014,Lerman2002,Berman2007,Galstyan2005,Sugawara1997}
time has been developed~\citep{Lerman2002,Berman2007,Galstyan2005,Sugawara1997},
which sidesteps the difficulty in modeling the average behavior of the swarm, by
instead modeling the \emph{change} in the average behavior of the swarm, which
is much easier.  It uses (1) differential equations to model the behavior of the
\emph{average} number of robots in $\TheSwarm$ in a given state, (2) discrete
difference equations to model the stochastic transitions between robot states,
and (3) stochastic simulation of discrete difference equations to compute state
transition rates for all robots. It draws on implementations of the stochastic
master equation in chemistry and statistical
physics~\citep{VanKampen2007}. Through the usage of rate constants and
population fractions in each state, it is possible to assess the general
behavior of a system under a variety of stochastic circumstances. Most
importantly, this approach has predictive control and performance
guarantees---crucial components to translating laboratory models into viable
real-world solutions without needing behavioral characterization of robot swarms
via simulation experiments.

Models in this differential equation paradigm operate on both the forward
problem, i.e., predicting collective behavior from features of the control
algorithm each robot runs~\citep{Lerman2002}, and the inverse problem, i.e.,
incorporating design constraints into algorithm design in order to produce a
desired collective behavior~\citep{Berman2007,Hsieh2008}. However, the models
are not overly robust, for two reasons. First, they make large simplifying
assumptions such as homogeneous agent distributions, homogeneous environments
(e.g., no obstacles and/or a completely visible arena), and Markov/semi-Markov
scenarios~\citep{Berman2007}.  Second, they require many free parameters and
extensive post-hoc model fitting which are specific to the implementation of the
algorithm under study. Nevertheless, many notable applications of various forms
of this methodology have appeared in the literature, demonstrating its practical
utility. A few noteworthy examples include the stick pulling
experiment~\citep{Ijspeert2001}, foraging of green/red pucks using agent
memory~\citep{Lerman2003a}, the ``house hunting
model''~\citep{Hsieh2008,Berman2007}, and ant-inspired models that collaborate
with or without communication~\citep{Sugawara1997}.

% CMM: It feels like you need a transition sentence at the end of this paragraph.
% It seems like it just drops off to me.

%%%%%%%%%%%%%%%%%%%%%%%%%%%%%%%%%%%%%%%%%%%%%%%%%%%%%%%%%%%%%%%%%%%%%%%%%%%%%%%%
% Background
%%%%%%%%%%%%%%%%%%%%%%%%%%%%%%%%%%%%%%%%%%%%%%%%%%%%%%%%%%%%%%%%%%%%%%%%%%%%%%%%
%
\section{Background in ODE modeling and diffusion}\label{sec:background}
\begin{figure}[ht]
  \centering
  \includegraphics[width=\linewidth]{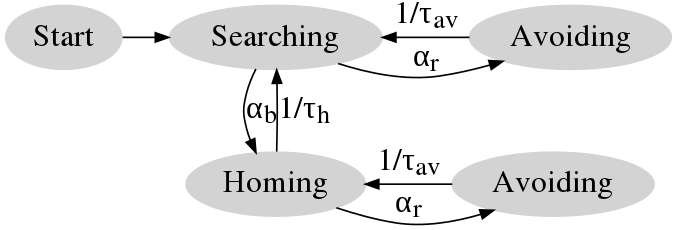}
  \caption{\label{fig:fsm} State diagram for a single robot. The \emph{Avoiding}
    state is duplicated to uniquely identify the collision avoidance context:
    avoidance while \emph{Homing} or avoidance while
    \emph{Searching}. Transition rates are described
    in~\cref{tab:ode-terms}. Note that the inverse of the amount of time a robot
    spends in a given state (e.g., $\tau_{h}$ for \emph{Homing}) is the rate of
    robots leaving the state. Low level details such as sensing and actuation
    are omitted}
\end{figure}

%
% [JRH]: Address ijcai#63 comment about being decoupled from real-robot
% contexts.
To model swarm collective behavior using ODEs, we consider the individual robot
Finite State Machine (FSM) (\cref{fig:fsm}).  Our FSM is identical to previous
work~\citep{Lerman2001,Lerman2002}. Each state maps directly to a single robot
behavior or a set/sequence of robot behaviors which together make up the robot
controller for executing a foraging task. It is a coarse-grained model of robot
behavior, which omits controller details such as sensing and actuation and
contains the \emph{minimum} number of states needed to describe the system
dynamics for the problem.  Such an approach is generally more mathematically
tractable.  If the results of the analysis do not sufficiently agree with the
observed collective behavior, other states can be added~\citep{Lerman2002}.

% [JRH] Address ijcai#63 comment about the method to go from FSM->ODE
% being case-specific.

Another benefit of the ODE modeling approach is that each of the states
in~\cref{fig:fsm} maps directly to an ODE describing this change, where the
transition rates for states become the ODE terms, and all ODEs can therefore be
directly written down from robot control algorithm characteristics.  We refer
the reader to~\citep{Lerman2001,VanKampen2007} for the theoretical underpinnings
of the validity of this translation, which can be applied to any SR system, not
just those doing a foraging task. The quantities modeled in~\cref{fig:fsm} are
listed in~\cref{tab:ode-terms}, and the ODE model of these quantities from
previous work~\citep{Lerman2001,Lerman2002} is summarized
in~\cref{eqn:lerman-ode-searching,eqn:lerman-ode-homing,eqn:lerman-ode-avoiding,eqn:lerman-ode-blocks}.
In our model, shown later in Section~\ref{ssec:method-ode}, we simplify the
original equations and replace some parameters with mathematical derivations.
\begin{table}[ht]
  \centering
  \begin{tabularx}{\linewidth}{ p{0.9cm} X }\hline
    {Quantity} &  {Description}  \\
    \hline
    $\alpha_r$ &   {Robot encounter rate of robots \emph{anywhere} in the arena} \\ [1ex]
    $\alpha_{r'}$\string^ & Robot encounter rate \emph{near} the nest \\[1ex]
    $\alpha_r$\string^ &   {Robot encounter rate of robots \emph{far away} from the nest} \\ [1ex]
    $\alpha_b$ &   {Robot encounter rate of blocks} \\ [1ex]
    $\tau_h$ &   {Mean robot homing time} \\ [1ex]
    $\tau_{av}$ &   {Mean robot time spent avoiding collision, per occurrence} \\ [1ex]
    $\SwarmNHoming$ & Mean number of robots returning to nest with blocks \\[1ex]
    $\SwarmNSearching$ & Mean number of robots searching for blocks\\[1ex]
    $\SwarmNAvoidingWhileHoming$ & Mean number of robots avoiding collision while homing\\ [1ex]
    $\SwarmNAvoidingWhileSearching$ & Mean number of robots avoiding collision while searching\\ [1ex]
    $\NBlocksInEnv{t}$\string^ & Mean number of blocks in the arena\\[1ex]
    $\SubAreaNBlocks{t}$* & Mean number of blocks in area $j$ of the arena\\[1ex]
    \hline
  \end{tabularx}\caption{\label{tab:ode-terms}Summary of ODE model
      components. Components
      with a \string^ are only in previous work, the one  with a * is only in our model.}
\end{table}

% [JRH] Add more explanation for ODE terms in original model to address IJCAI
% reviewer#7 comment.
\cref{eqn:lerman-ode-searching} describes the change in the number of robots in
the searching state, which decreases as searching robots pickup blocks or
encounter other robots and switch to collision avoidance.
\begin{flalign}
  \frac{d\SwarmNSearching}{dt}
  =~&-\alpha_b\SwarmNSearching\big[\NBlocksInEnv{t} - \SwarmNHoming -\SwarmNAvoidingWhileHoming\big] \label{eqn:lerman-ode-searching} \\
  &-  \nonumber \alpha_r\SwarmNSearching\big[\SwarmNSearching + \TheSwarmSize\big]\\
  &+ \nonumber \frac{1}{\tau_h}\SwarmNHoming + \frac{1}{\tau_{av}}\SwarmNAvoidingWhileSearching
\end{flalign}
The rate at which robots leave the searching state and switch to collision
avoidance, $\alpha_{r}\SwarmNSearching[\SwarmNSearching + \TheSwarmSize]$, can
be understood as follows. When a searching robot encounters another searching
robot, both switch to the collision avoidance state, decreasing the
$\SwarmNSearching$ by two; when a searching robot encounters either a homing
robot or a robot already in the collision avoidance state, $\SwarmNSearching$
decreases by one. The total decrease is then proportional to $2\SwarmNSearching
+ \SwarmNHoming + \SwarmNAvoidingWhileSearching + \SwarmNAvoidingWhileHoming =
\SwarmNSearching + \TheSwarmSize$, where $\TheSwarmSize$ is the total number of
robots in the swarm. Finally, searching robots encounter other robots at rate
$\alpha_r$, regardless of the state of the other robot.
\cref{eqn:lerman-ode-searching} increases as homing robots deposit blocks in the
nest or as searching robots exit the collision avoidance state.

\cref{eqn:lerman-ode-homing} describes the change in the number of robots in the
homing state, which increases as robots acquire and pickup blocks or leave the
collision avoidance state, and decreases as robots enter the
collision avoidance state or deposit their block in the nest.
\begin{flalign}
  \frac{d\SwarmNHoming}{dt} =~&
  \alpha_b\SwarmNSearching\big[\NBlocksInEnv{t} - \SwarmNHoming -\SwarmNAvoidingWhileHoming\big] \label{eqn:lerman-ode-homing} \\
  &- \nonumber \alpha_{r'}\SwarmNHoming\big[\SwarmNHoming + \TheSwarmSize\big]\\
  & \nonumber  - \frac{1}{\tau_h}\SwarmNHoming + \frac{1}{\tau_{av}}\SwarmNAvoidingWhileHoming
\end{flalign}
The term $\alpha_{r'}\SwarmNSearching[\SwarmNHoming + \TheSwarmSize]$ can be
understood analogously to the one in~\cref{eqn:lerman-ode-searching}. The model
assumes that the rate at which homing robots encounter other robots is different
than for searching robots, reasoning that there will be more congestion near the
nest, and therefore uses a separate parameter $\alpha_{r'}$ to account for this.

\cref{eqn:lerman-ode-avoiding} describes the change in the number of
robots avoiding collision with other robots.
\begin{flalign}
  \frac{d\SwarmNAvoidingWhileSearching}{dt} =~& \alpha_{r'}\SwarmNHoming\big[\SwarmNHoming +
  \TheSwarmSize\big] - \frac{1}{\tau_{av}}\SwarmNAvoidingWhileSearching \label{eqn:lerman-ode-avoiding}
\end{flalign}

\cref{eqn:lerman-ode-blocks} describes the change in the number of free blocks
available for robots to find; it decreases whenever a searching robot acquires a
block (i.e., once deposited in the nest, blocks are not re-distributed in the
arena).
\begin{flalign}
  \frac{d\NBlocksInEnv{t}}{dt} =~&-\frac{1}{\tau_h}\SwarmNHoming\label{eqn:lerman-ode-blocks}
\end{flalign}

In addition to using ODEs to model collective behavior, we also draw on
diffusion theory.  In diffusing systems composed of homogeneous particles
undergoing \emph{normal diffusion}, the average particle displacement is
proportional to the diffusion time, i.e., a linear relationship; this assumes
the particle is moving in an infinite, structureless medium close to
equilibrium.  However, there are important circumstances in which the
relationship between average particle displacement and time is non-linear,
referred to as \emph{anomalous
  diffusion}~\citep{Olivera2019,Metzler2014,vlahos2008normal}. For instance, in
biological systems, interactions with other particles or membranes could
influence the observed diffusion for macro-proteins as they traverse through
biological media~\citep{Holek2009,Weiss2003,Nicolau2007}. Anomalous subdiffusion
arises due to crowding in a concentrated system, which can make it heterogeneous
and disordered~\citep{Ghosh2015}. Crowded environments can obstruct the path or
make certain spaces inaccessible to the diffusing particle and prompt the rise
of this behavior.

%%%%%%%%%%%%%%%%%%%%%%%%%%%%%%%%%%%%%%%%%%%%%%%%%%%%%%%%%%%%%%%%%%%%%%%%%%%%%%%%
% Model
%%%%%%%%%%%%%%%%%%%%%%%%%%%%%%%%%%%%%%%%%%%%%%%%%%%%%%%%%%%%%%%%%%%%%%%%%%%%%%%%
%%% Local Variables:
%%% mode: latex
%%% TeX-master: "2021-ar-main"
%%% End:
\section{Our Generalized ODE Foraging Model}\label{sec:method}
%
%------------------------------------
\subsection{Changes to previous work}\label{ssec:method-improvements}
The ODE model described
in~\cref{eqn:lerman-ode-searching,eqn:lerman-ode-homing,eqn:lerman-ode-avoiding,eqn:lerman-ode-blocks}
from~\citep{Lerman2002} has some limitations which we address in our model.

\subsubsection{Uniformity assumptions}

\begin{figure}[ht]
  {\includegraphics[width=0.99\linewidth,height=0.15\textwidth]{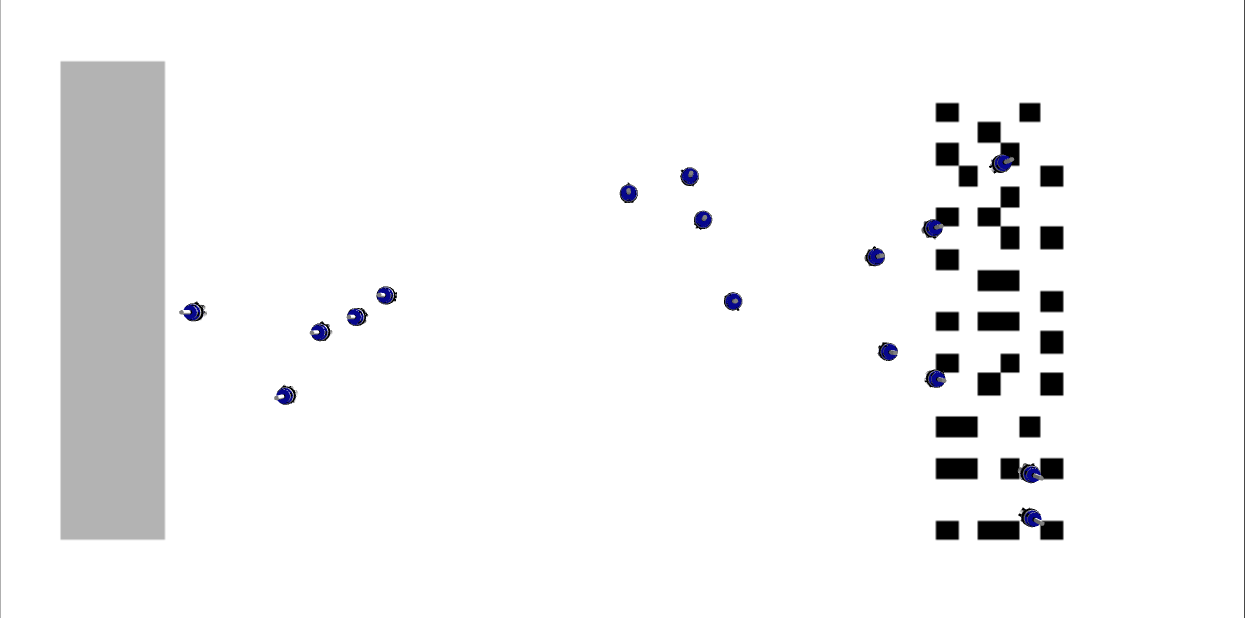}}
  \footnotesize{(a) \emph{Single source} (SS) object distribution, with objects on the right
  and the nest on the left}\rule[-1ex]{0ex}{1ex}
  \newline
  \includegraphics[width=0.99\linewidth,height=0.15\textwidth]{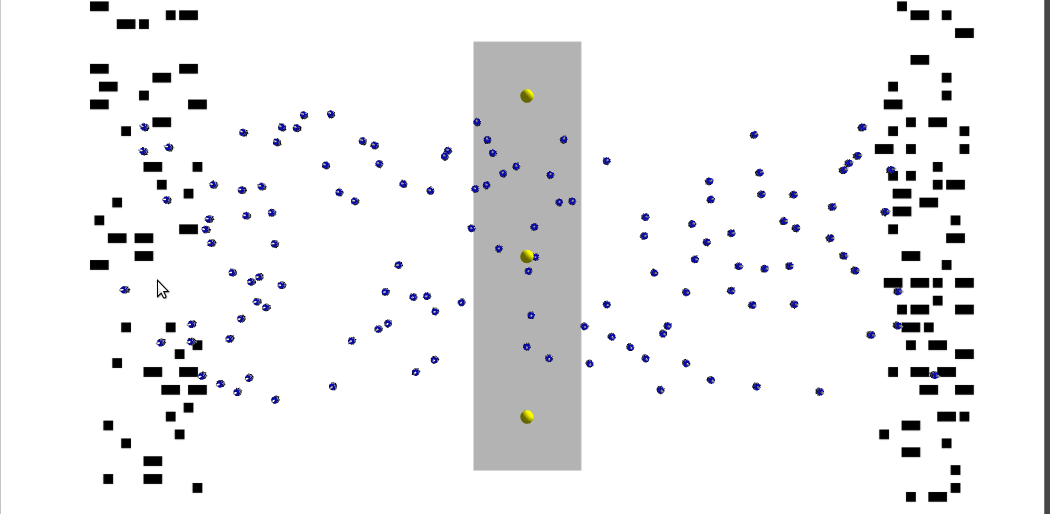}
  \footnotesize{(b) \emph{Dual source} (DS) object distribution, with
  objects on the right and left and the nest in the center\rule[-1ex]{0ex}{1ex}}
  \newline
\begin{minipage}[t]{.48\linewidth}
  {\includegraphics[width=.99\linewidth]{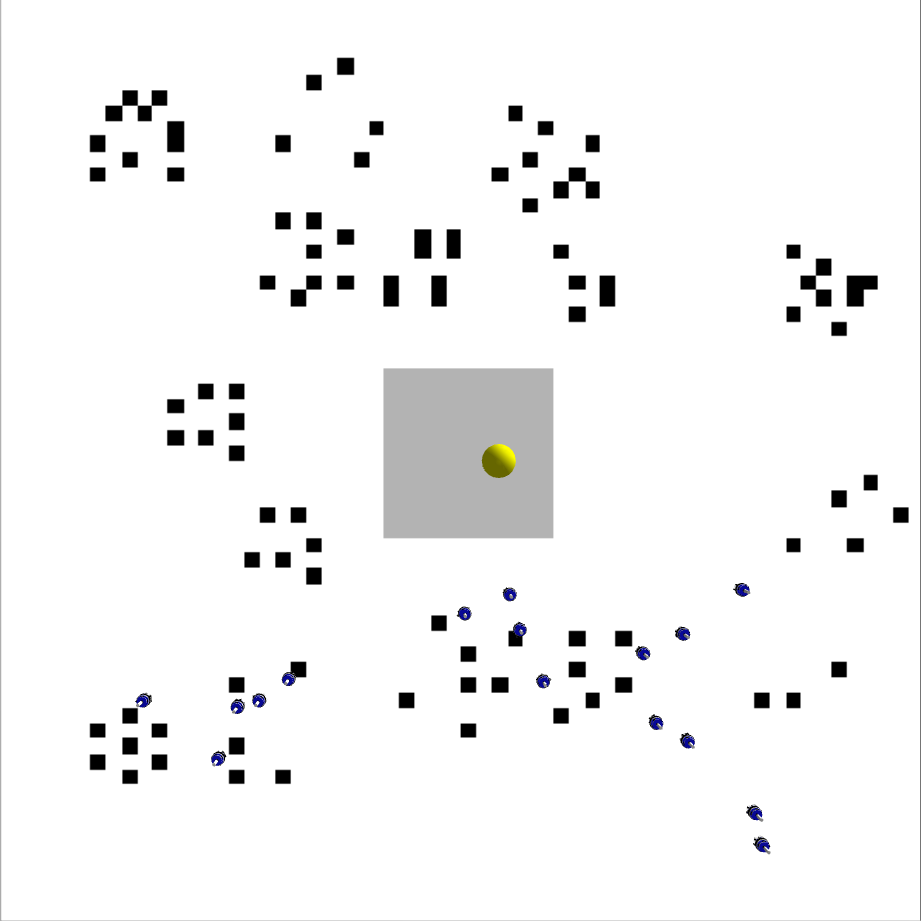}}
  \footnotesize{(c) \emph{Power law } (PL) object distribution, with the nest in the center\rule[-1ex]{0ex}{1ex}}
  \end{minipage}\hfill
\begin{minipage}[t]{.48\linewidth}
  {\includegraphics[width=.99\linewidth]{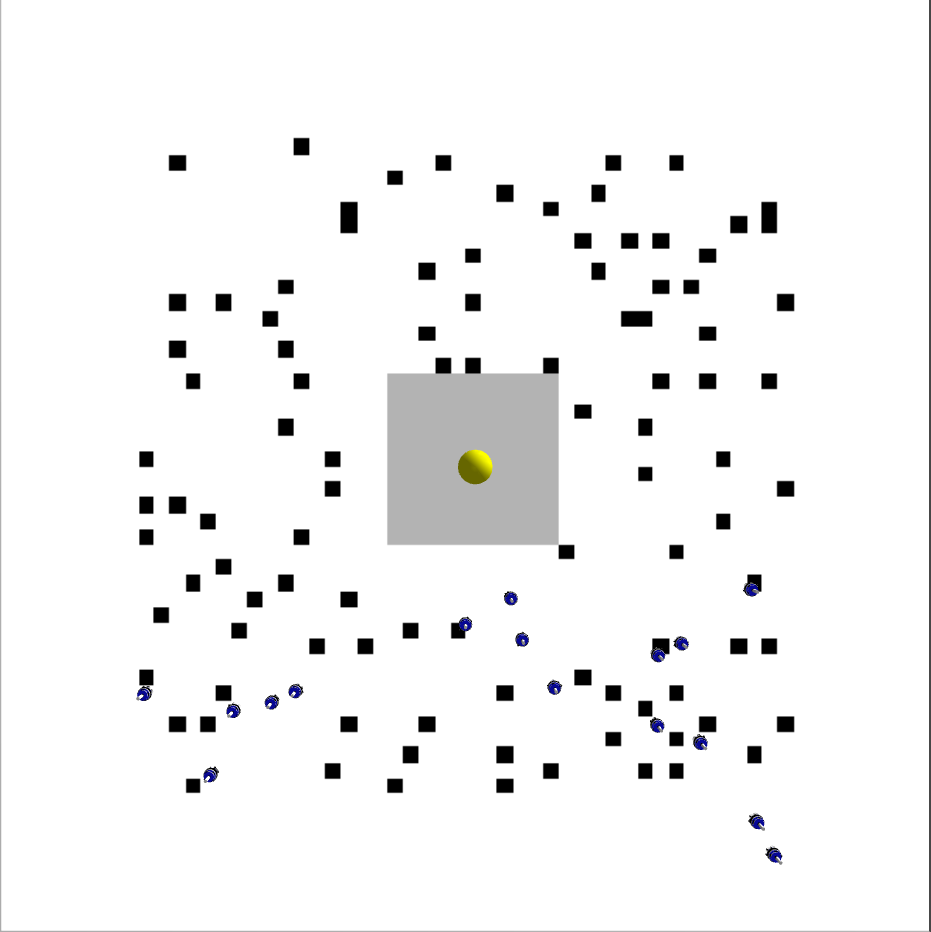}}
  \footnotesize{(d) \emph{Random} (RN) object distribution, with the nest in the center}
  \end{minipage}
  \caption{Examples of distributions of objects for foraging}\label{fig:foraging-dists}
\end{figure}

In previous work $\TheSwarm$ is assumed to be uniformly distributed in 2D space,
and blocks to be scattered randomly; therefore, the model is only accurate on
scenarios meeting these criteria. In \cref{fig:foraging-dists} we show some of
those cases.  \emph{Random} (RN) block
distributions~\citep{Sugawara1997,Hecker2015} (\cref{fig:foraging-dists}d), are
appropriate in scenarios such as order fulfillment in a warehouse, but many
scenarios cannot be modeled by such
distributions~\citep{Campo2007,Sugawara1997,Castello2016,Hecker2015,Lerman2001}.
For example, transferring material from one side of a building to another
location requires a \emph{single source} (SS) (\cref{fig:foraging-dists}a) or
\emph{dual source} (DS) (\cref{fig:foraging-dists}b) block distribution model;
such scenarios are studied
in~\citep{Harwell2019a,Ferrante2015,Pini2011b,Harwell2020a}.
For other scenarios, such as evacuation of civilians from a disaster zone, the
block distribution cannot be inferred \emph{a priori}, and a \emph{power law}
(PL) distribution (\cref{fig:foraging-dists}c), in which blocks are clustered in
groups of various sizes, is appropriate~\citep{Hecker2015,Harwell2020a}.  By
considering the distribution of blocks in a general way using spatial clusters,
we guarantee that the model itself describes underlying characteristics of the
environment and is robust enough to be applied to a wide range of foraging
scenarios.

\subsubsection{Invalid steady-state assumptions }
The model assumes that $\TheSwarm$ has reached steady-state, but at $t=0$
assumes a finite number of objects for the swarm to find
(\cref{eqn:lerman-ode-blocks}). This does not model accurately long-running SR
systems or those with a small number of blocks relative to $\TheSwarmSize$.  We
formulate our ODE model to handle both steady-state and non steady-state
behaviors by considering two types of operating environments. First, those in
which any block deposited in the nest is re-distributed immediately in the
operating area, so the overall number of blocks available for robots to find in
the arena remains the same over time. Second, those in which any block deposited
in the nest is not re-distributed, resulting in fewer blocks available for
robots to find over time, which is the paradigm used in~\citep{Lerman2002}. Our
model treats non steady-state foraging environments as a special case of
steady-state behaviors by setting to zero the number of areas where blocks can
be distributed after nest deposition (see $A_j$ later
in~\cref{eqn:ode-subarea-blocks}). In this work, we are interested in
steady-state behaviors.

\subsubsection{Many free parameters}
Free parameters do not invalidate the theoretical basis of the model, but limit
its reuse by requiring iterative parameter refinement. The need to do post-hoc
model fitting requires experimental simulations, which makes it difficult to use
the model as a predictive tool. We instead derive analytical expressions for all
parameters except $\tau_{av}$ from scenario characteristics, greatly reducing
the need for post-hoc model fitting and reducing the number of free parameters
from five to one. $\tau_{av}$ is directly computable from the robot controller
characteristics, or otherwise obtainable via experiments with a single robot. We
introduce $\ScenarioCACharacterization{\TheScenario}$, to characterize the
inter-robot interference in scenario $\TheScenario$. Although it is a free
parameter, it has a strong semantic context, making fitted values more
meaningful and suggesting the possibility of future derivation.

%------------------------------------
\subsection{Preliminaries}\label{ssec:method-preliminaries}

To make accurate predictions of collective swarm behavior in the steady-state
for each of the block distributions shown in~\cref{fig:foraging-dists}, our
model makes the following assumptions:
\begin{enumerate}\setlength{\itemsep}{1pt}\setlength{\parskip}{0pt}\setlength{\parsep}{0pt}

  % [JRH] ijcai#10 did not like this assumption, so I did additional
  % experiments with variable density AND across scales to address their
  % concern.
\item

  The swarm density $\SwarmDensity$ is relatively low, so the behavior of
  $\TheSwarmSize$ robots can be approximated by a linear function of the
  behavior of a single robot. An important consequence of this is that we can
  ignore the effect of a robot avoiding collision and encountering
  \emph{another} robot during its avoidance maneuvers. This is not necessarily a
  limiting assumption, as many real-world applications require sparsely
  distributed swarms~\citep{Tarapore2020}. We relax it as part of our model
  evaluation experiments in~\cref{sec:results}.

\item

  $\TheSwarmSize$ is sufficiently large such that $\TheSwarm$ can be
  approximated using the mean field model. We relax also this assumption as part
  of our model evaluation experiments~\cref{sec:results}.

\item

  $\TheSwarm$ is homogeneous, and the control algorithm used by the robots is a
  Correlated Random Walk (CRW)~\citep{Renshaw1981,Harwell2020a}, which is a
  random walk where the direction of the next step is biased based on the
  direction of the previous step.  The bias angle $\theta$ is drawn from a
  probability distribution $f(\theta)$~\citep{Codling2010}.  Robots do a CRW at
  an average velocity ${v_s}$ until they acquire a block, which they then
  transport to the nest using phototaxis, i.e., motion in response to light, at
  a constant speed ${v_h}$. Robots have no memory.

\item

  To reduce congestion in the nest, robots do not return to the exact nest
  center to drop their carried object, i.e., they employ a reactive avoidance
  strategy. Instead, they choose a random point along their trajectory from
  where they enter the nest to treat as the nest center. This shortens the
  homing distance, and can be computed as shown later
  in~\cref{ssec:method-tau-h}.

\item

    $\TheSwarm$ has reached steady-state at some $t \gg 0$. This maps naturally
    to swarms with long-running autonomy, such as those deployed on large-scale
    agriculture and industrial automation applications.

\end{enumerate}

None of the assumptions we made reduces the utility of our model in its
application to real-world problems; i.e., they are consistent with the
constraints imposed by the problems themselves and commonly used robot hardware.
We will relax some of them during the evaluation of our model, showing that
while they are helpful during derivation, they are not essential for practical
applications of our model.

%------------------------------------
\subsection{ODE model}\label{ssec:method-ode}
Let $\AllScenarios = \{SS,DS,PL,RN\}$ be the set of scenarios based on the block
distributions shown in~\cref{fig:foraging-dists}. In each
$\TheScenario\in{\AllScenarios}$, let the area where blocks can be distributed
be a subset of the overall area $\OperatingArea$ of the arena. Let
$j=1,\dots,~J$ be the sub-areas within $\OperatingArea$ in which blocks can be
distributed, each described by a tuple
$(A_j,\SubAreaCenter,\mathbf{d}_j,\SubAreaBlockDensity)$. $A_j$ is the area
occupied by the sub-area $j$, $\SubAreaCenter$ is the center of the sub-area,
$\mathbf{d}_j$ is its dimensions, and $\SubAreaBlockDensity$ is the mean
steady-state block density within the sub-area. The area within $\OperatingArea$
where blocks can be found is the union of these disjoint subsets:
$\OperatingArea_d=\cup{A_j}$.

% [JRH] Address ijcai#7 comment about our block distribution model not
% being heterogenous/assumes encounter rate is the same for all clusters in the
% arena.

The value of $\SubAreaBlockDensity$ varies across sub-areas in our model, so
that the block encounter rate $\alpha_b$ can be captured accurately even in
extreme non-homogeneous block distributions. As an example, consider a
degenerate case in which there are two areas, both represented by a circular
sector around the nest. Robots would likely forage only from the inner circle
and rarely reach the outer circle, and so a single $\SubAreaBlockDensity$ is not
sufficient to capture collective dynamics in such cases. We require both
$\SubAreaBlockDensity$ and the robot spatial occupancy distribution in order to
accurately compute $\alpha_b$ in a general way (see \cref{ssec:method-alpha-b}).

\cref{eqn:ode-searching,eqn:ode-homing,eqn:ode-avoiding} describe our
generalized ODE model.  We simplified the original equations
from~\citep{Lerman2002,Lerman2001}, by removing $\alpha_r'$ and replaced
$\tau_h,\alpha_b,\alpha_r$ with mathematical derivations.  The interpretation
of~\cref{eqn:ode-searching,eqn:ode-homing,eqn:ode-avoiding} is the same as
\cref{eqn:lerman-ode-searching,eqn:lerman-ode-homing,eqn:lerman-ode-avoiding} in
previous work described earlier in~\cref{sec:background}.
\begin{flalign}
  \frac{d\SwarmNSearching}{dt}
  =~&-\alpha_b - \alpha_r + \ \frac{1}{\tau_h}\SwarmNHoming + \frac{1}{\tau_{av}}\SwarmNAvoidingWhileSearching\label{eqn:ode-searching} \\
  \frac{d\SwarmNHoming}{dt} =~& \alpha_b - \alpha_r - \frac{1}{\tau_h}\SwarmNHoming + \frac{1}{\tau_{av}}\SwarmNAvoidingWhileHoming\label{eqn:ode-homing} \\
  \frac{d\SwarmNAvoidingWhileSearching}{dt} =~& \alpha_r - \frac{1}{\tau_{av}}\SwarmNAvoidingWhileSearching\label{eqn:ode-avoiding}
\end{flalign}
\cref{eqn:ode-subarea-blocks} uses the described block modeling method to
capture the underlying behavioral dynamics of the swarm. We made the following
additional assumptions about the block distribution for its derivation. First,
whenever a block is redistributed all $j$ are selected with probability
proportional to the fraction of distributable area they contain. Second, blocks
are distributed uniformly within a given $j$. Third, every $j$ can hold any
number of blocks, allowing two blocks to occupy the same location (i.e.,
stacking).  We note that under our-steady state
assumption,~\cref{eqn:ode-subarea-blocks} can be solved analytically given
$\NBlocksInEnv{0}$ and does not need to be solved numerically. We also observe
that from~\cref{eqn:ode-subarea-blocks} the number of blocks in the arena
available for robots to find can increase \emph{and} decrease, addressing one of
the weaknesses in the model from previous work.
\begin{align}
  \frac{d\SubAreaNBlocks{t}}{dt} =~&\frac{1}{\tau_h}\SwarmNHoming\frac{A_j}{\OperatingArea_d} - \alpha_b\frac{A_j}{\OperatingArea_d}~~{j}=1,\ldots,~{J}\label{eqn:ode-subarea-blocks}
\end{align}
Next we derive analytical models for $\tau_h,\alpha_b$, and $\alpha_r$ from the
arena geometry, number of blocks, block distribution, etc. We do not derive
$\tau_{av}$, because it depends intrinsically on the interference avoidance
strategy employed by $\TheSwarm$ and therefore cannot be derived independently
from $\kappa$ without additional assumptions.

%------------------------------------
\subsection{Derivation of homing time $\tau_h$}\label{ssec:method-tau-h}
%
% [JRH] Update this section to be clearer, addressing ijcai#7 comment.
%
We build the intuition behind the block acquisition probability for a robot at
location $\ArenaXY$ in the arena as follows, modeling the nest as a single point
$\NestXY$.  Since searching begins from the nest, the density of
$\SwarmNSearching$ must be \emph{greater} near the nest because robots perform
\emph{biased} random walks which originate from a common point. Consequently,
this non-uniform swarm spatial distribution means that the mean distance from
the nest at which a searching robot encounters a block is \emph{not} the same as
the mean distance of a block from the nest. From~\citep{Codling2010}, we have
that the spatial occupancy distribution from the central point, as a result of a
biased random walk with bias distribution $Uniform(-\theta,\theta)$, falls off
linearly, and we would intuitively expect the following:
%
% [JRH] Try to justify the decisions in the heuristic density function better,
% addressing ijcai#10 comment.
\begin{enumerate}
\item

    It is moderated by $\SubAreaBlockDensity=\SubAreaNBlocks{t}/A_j$, because
    the rate of decay of the mean block acquisition distance as a function of
    distance from the nest within a given $A_j$ is slower for low
    $\SubAreaBlockDensity$. Consequently, a $j$ sparsely populated with blocks
    will have minimal effect on the overall swarm block acquisition probability
    distribution, while a $j$ densely packed with blocks will create an area of
    higher acquisition probability.

\item

    $\SubAreaBlockDensity$ would play an exponentially moderating role only when
    the block acquisition location $\ArenaXY$ is close to $\NestXY$, such as for
    RN or PL block distributions. For SS and DS distributions, where the mean
    distance from a block to the nest is large, the effect of
    $\SubAreaBlockDensity$ on block acquisition locations should be minimal.
\end{enumerate}
Using the results of~\citep{Codling2010} and these intuitions, we formulate
\cref{eqn:acq-pdf-single} as a close approximation of the occupancy distribution
of a single random walker performing a correlated random walk starting from a
central point $\NestXY$. $C$ is a normalization constant to ensure
$p_{acq_{j}}(\ArenaXY)$ integrates to $1$ over all $j$.
\begin{equation}\label{eqn:acq-pdf-single}
  p_{acq_{j}}(\ArenaXY) = \frac{C}{\big({\sqrt{\dtwonorm{\ArenaXY - \NestXY}} -\frac{\ln{(\SubAreaBlockDensity)}}{2\SubAreaBlockDensity}}\big)^2}
\end{equation}
Having defined the probability density function, we now derive the expected
acquisition location by finding the expected values of the marginal density
functions in $x$ by integrating~\cref{eqn:acq-pdf-single} and summing over all
$j$:
\begin{equation}
  \ExpectedAcqX = \sum_{J}\int_{x}\int_{y}p_{acq_{j}}(\ArenaXY)x{dx}{dy}
\end{equation}
and similarly for $y$. We now write an expression for $\tau_h^1$:
\begin{equation}
  \tau_h^1 = \frac{\dtwonorm{\ExpectedAcqXY - \NestXY} - d_{cr}}{v_h}
\end{equation}
$v_h$ is the robot phototaxis velocity, specified in the input configuration,
and $\dtwonorm{\ExpectedAcqXY - \NestXY}$ is the expected distance an acquired
block will be from the center of the nest.
%
% [JRH] Provide details to make the employed congestion avoidance strategy
% clear, address ijcai#10 comment.
%
$d_{cr}$ is the distance the homing path is shortened due to the employed
congestion reduction strategy, and is straightforward to calculate from the
arena geometry.

We provide an example of computing $d_{cr}$ for a square arena, which in this
work corresponds to the RN and PL scenarios; similar calculations can be done
for the other arena shapes. Using a square nest of length $L$, without loss of
generality we find the average distance to the origin of a randomly selected
point in a region

$R\coloneqq \{(x,y) : 0 \le y \le x \le L/2\}$ that is a
triangle 1/8 of the nest. The uniform density on this region is then:
\begin{equation}
  f(x,y) =
  \begin{cases}
    8/L^2 & \text{if}~(x,y) \in R \\
    0 & \text{otherwise}
  \end{cases}
\end{equation}
Then, the average distance $d_{cr}$ is computed via:
\begin{equation}
  \begin{aligned}
    d_{cr} &= \frac{8}{L^2}\int_{0}^{L/2}\int_{0}^x{\sqrt{x^2 + y^2}}dydx \\
    &= \frac{L}{6}(\sqrt{2} + \ln(1 + \sqrt{2}))
  \end{aligned}
\end{equation}

Finally, to derive $\tau_h$, we note that under our assumption of low to
moderate $\SwarmDensity$, the homing time increases linearly with
$\TheSwarmSize$ according to the expected value of time lost due to inter-robot
interference~\citep{Lerman2002}, averaged across all robots:
\begin{equation}\label{eqn:tau-h}
  \tau_h = \tau_h^1\big[1 + \frac{\alpha_r\tau_{av}}{\TheSwarmSize}\big]
\end{equation}

%------------------------------------
\subsection{Derivation of block acquisition rate $\alpha_b$}\label{ssec:method-alpha-b}
Using our mean-field assumption, $\TheSwarm$ can be approximated as a fluid
composed of robot particles, and be considered to obey many of the same laws; in
the long-term limit, the governing equation for the biased random walk used in
this work is the advection-diffusion equation~\citep{Codling2010}. Given
sufficiently simple robots, this approximation gives good
results~\citep{Codling2010,Pang2018}. Using simple robots does
not detract from the utility of the system, which mimics the
self-organizing behavior of real swarm populations such as social insects.

Using this intuition, we obtain $\alpha_b$ by computing the mean time it takes a
robot ``particle'' starting at $\NestXY$ to ``diffuse'' within the operating
area $\OperatingArea$ to the expected acquisition location
$\ExpectedAcqXY$. Viewing $\dtwonorm{\ExpectedAcqXY - \NestXY}$ as the Root Mean
Square (RMS) displacement distance and assuming a linear relationship between
displacement and diffusion time, we obtain:
\begin{equation}\label{eqn:block-acq-rate}
  \frac{1}{\alpha_b} = \frac{{\dtwonorm{\ExpectedAcqXY - \NestXY}}^2}{2\CRWDiffusionD{\TheSwarmSize}}
\end{equation}
%
% [JRH] Emphasize that alpha_b DOES consider multiple block distributions,
% densities, etc., addressing ICJAI reviewer#10 comment.
%
where $\CRWDiffusionD{\TheSwarmSize}$ is the diffusion constant for a swarm of
$\TheSwarmSize$ robots, and $\alpha_b$ is the expected time to diffuse from
$\NestXY$ to $\ExpectedAcqXY$. We note that while the calculation for $\alpha_b$
does not directly consider the distribution of blocks in the arena and their
densities within a given block cluster, it depends on $\ExpectedAcqXY$, which
does.

An exact calculation of $\CRWDiffusionD{\TheSwarmSize}$ is out of the scope of
this paper, so we approximate it as shown in~\cref{eqn:crw-diffusion} using the
results of the RMS diffusion for CRW~\citep{Codling2010} and our intuition that
smaller $\theta$ will result in quicker swarm diffusion (more straight line
motion):
\begin{equation}\label{eqn:crw-diffusion}
  \CRWDiffusionD{\TheSwarmSize} = \TheSwarmSize\frac{\ScenarioHeterogeneity{\TheScenario}\DiffusionXY}{\DiffusionTheta}
\end{equation}
where $\ScenarioHeterogeneity{\TheScenario}$ is a per-scenario
parameter characterizing the linearity of the diffusion
rate, and $\DiffusionXY$ is defined as follows from~\citep{Codling2010}:
\begin{equation}\label{eqn:diffusion-xy}
  \DiffusionXY = \frac{{{v_s}}^2}{4t}\underbrace{\int_{-\pi}^{\pi}(1 \pm \cos{2\theta})f(\theta)d\theta}_{\text{$\DiffusionTheta$}}
\end{equation}
where ${v_s}$ is the robot searching velocity and a scenario parameter.

Under ``normal'' circumstances with inert
particles,~\cref{eqn:block-acq-rate,eqn:crw-diffusion} give good results because
system diffusion varies linearly with time, which these equations
require. However, many foraging environments are heterogeneous and have
non-uniform distributions of blocks. As a result, the swarm may experience
non-linear \emph{anomalous diffusion}~\citep{Hasnain2018,Woringer}, i.e.,
crowding in some areas of the environment, giving rise to the need for
$\ScenarioHeterogeneity{\TheScenario}$, which implicitly captures these
artifacts, as shown in~\cref{eqn:crw-diffusion}.

% [JRH] Can be cut for space. I don't actually think this provides any new
% information...

%% \begin{figure}[t]
%%   {\includegraphics[width=0.9\linewidth]{figures/updated_hetero.png}}
%%   \caption{\label{fig:hetero-plot-figure}\footnotesize{Scenario
%%       heterogeneity. $\ScenarioHeterogeneity{\TheScenario}$
%%       in~\cref{eqn:scenario-heterogeneity}, we have restored the linear
%%       relationship between diffusion time and Mean Squared Displacement (MSD).}}
%% \end{figure}

To calculate $\ScenarioHeterogeneity{\TheScenario}$ from first principles, we
take the following approach. First, we calculate the environment heterogeneity
$\EnvHeterogeneity{\TheScenario}$ in reference to an environment with perfectly
uniform block distribution (homogeneous case). For the square RN and PL
scenarios, we compute the variance of the average inter-cluster distance.  For
the rectangular SS and DS scenarios with $J \le{2}$, the average inter-cluster
distance is not meaningful, so we calculate $\EnvHeterogeneity{\TheScenario}$ as
the mean cluster distance from the ``ideal'', which is defined as an environment
in which all $\SubAreaCenter$ are incident with $\NestXY$. Modulating
$\EnvHeterogeneity{\TheScenario}$
in~\cref{eqn:env-variance-sq,eqn:env-variance-rect} by the expected level of
crowdedness within each cluster we obtain an expression for
$\ScenarioHeterogeneity{\TheScenario}$ which accounts for the number of block
clusters ($J$) and the area cumulatively covered by all clusters
($\OperatingArea_d$):
\begin{equation}\label{eqn:env-variance-sq}
  \ScenarioHeterogeneity{\TheScenario} = \underbrace{\frac{1}{J}\sum\limits_J{E(\SubAreaCenter - E[\SubAreaCenter]) (\SubAreaCenter - E[\SubAreaCenter])^T}}_{\EnvHeterogeneity{\TheScenario}}\frac{\OperatingArea_d J}{\NBlocksInEnv{0}}
\end{equation}
\begin{equation}\label{eqn:env-variance-rect}
\ScenarioHeterogeneity{\TheScenario} = \underbrace{\frac{1}{J}\sum\limits_J{\dtwonorm{\SubAreaCenter - \NestXY}}}_{\EnvHeterogeneity{\TheScenario}}\frac{\OperatingArea_d J}{\NBlocksInEnv{0}}
\end{equation}

\begin{figure}[t]
  {\includegraphics[width=0.9\linewidth]{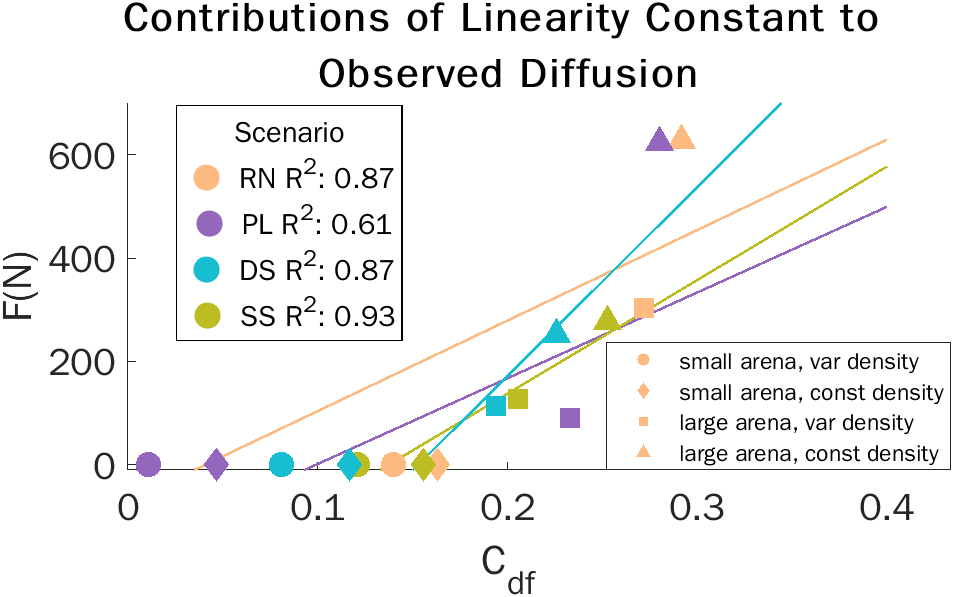}}
  \caption{\label{fig:diffusion-with-hetero}Observed linearity between the
    calculated diffusion constant $\ScenarioHeterogeneity{\TheScenario}$ for
    each scenario $\TheScenario$. We show four variants of each scenario: a
    small arena with small $\TheSwarmSize$ with constant and variable swarm
    density $\SwarmDensity$, and a large arena with large $\TheSwarmSize$ with
    constant/variable $\SwarmDensity$. These variants show our calculations for
    $\ScenarioHeterogeneity{\TheScenario}$ when (1) finite-size effects are
    present (small $\TheSwarmSize$), and when the mean-field approximation is
    valid (large $\TheSwarmSize$), and (2) under conditions favorable to linear
    modeling (constant $\SwarmDensity$ across $\TheSwarmSize$) and under
    conditions unfavorable to linear modeling (variable $\SwarmDensity$ across
    $\TheSwarmSize$). We see significant linear across all variants ($R^2 \ge
    0.67$).}
\end{figure}

In~\cref{fig:diffusion-with-hetero}, we plot the calculated values for
$\ScenarioHeterogeneity{\TheScenario}$ for all scenarios to verify we have
restored linearity of diffusion, and see strong correlations between
$\ScenarioHeterogeneity{\TheScenario}$ and $\CRWDiffusionD{\TheSwarmSize}$ for
all block distributions. As expected we see the highest heterogeneity associated
with the lowest diffusion rate (PL), and the lowest heterogeneity associated
with the highest diffusion rate (RN). With more observed heterogeneity within
the environment, the scaling factor necessary to restore the linearity of the
diffusion rate increases to accommodate for this occurrence. We also note that
the differing $x$ intercepts depend on the block distribution, and that as the
size of the arena increases the crowding observed is not as pronounced (less
anomalous diffusion). This might be due to the decreased likelihood of robots
being ``trapped''~\citep{Oleksandr} by obstacles in the environment, leaving the
robot more able to continue moving through the arena.

%% \cref{fig:diffusion-plot-figure} %Fig. 4.3 indicates how the calculated
%% $\ScenarioHeterogeneity$ depicts a high correlation between the observed
%% diffusion constants for most block distributions (DS, SS, RN).

%% Rather than a pure free parameter, the notion of subdiffusion is used as a
%% guiding principle that can guide the assignment of diffusion constants for
%% robots in different block distributions. The relationship between the
%% heterogeneity constants calculated and the diffusion constants used to
%% predict swarm behavior is represented in Fig. 4.2.

% [JRH] TODO: for final submission change 'density' in the legend to \rho_s to
% reduced crowdedness in the figure.

%------------------------------------
\subsection{Derivation of robot encounter rate $\alpha_r$}\label{ssec:method-alpha-r}
To derive $\alpha_r$, we use our assumption of low to moderate $\SwarmDensity$,
to model the robot encounter rate as a function of $\alpha_{r}^{1}$ and
$\CRWDiffusionD{\TheSwarmSize}$. We view \cref{fig:fsm} as a queueing network,
where robots are either performing collision avoidance maneuvers or not.  The
input/output transition rates for a state are summed to form the arrival and
service rates for the collision avoidance queue $\QueueCA$
($\lambda_{ca}=\alpha_{r}^{1}$ and $\mu_{ca}=1/\tau_{av}$, respectively).
Modeling $\QueueCA$ as a M/M/1 queue, i.e., at most one robot exits collision
avoidance per $\Delta{t}$, which is reasonable if $\Delta{t}$ is small, we can
write the following using Little's Law~\citep{Seda2017} as:
\begin{equation}\label{eqn:alpha_r}
  \alpha_{r} = \frac{\SwarmNAvoiding}{\tau_{av}} - \alpha_{r}^1\SwarmNAvoiding
\end{equation}
The second term in~\cref{eqn:alpha_r} is a corrective factor accounting for
robots that experience interference due to encountering arena walls, \emph{not}
other robots, which is simply the scaled rate at which a single robot
experiences interference.  We cannot use $\SwarmNAvoiding$ directly, as
$\alpha_{r}$ needs to be computed \emph{a priori}, but we can estimate it as a
function of $\alpha_{r}^{1}$ and $\SwarmNAvoidingOneRobot$, using our intuition
regarding swarm diffusion:
\begin{equation}\label{eqn:n-avoiding-est}
  \hat{\SwarmNAvoiding} = \SwarmNAvoidingOneRobot\frac{\CRWDiffusionD{\TheSwarmSize}}{\DiffusionTheta}\ScenarioCACharacterization{\TheScenario}
\end{equation}
where $\alpha_{r}^{1}$ and $\SwarmNAvoidingOneRobot$ can be computed from
$\kappa$ using the results of~\citep{Codling2010}. We increase the influence of
$\theta$ in~\cref{eqn:n-avoiding-est} by introducing another $\DiffusionTheta$
in the denominator; smaller $\theta$ will result in more inter-robot
interference due to straight line motion.
$\ScenarioCACharacterization{\TheScenario}$ characterizes the sub- or
super-linearity of~\cref{eqn:n-avoiding-est} for a scenario $\TheScenario$ in
relation to the random (RN) scenario.

%%%%%%%%%%%%%%%%%%%%%%%%%%%%%%%%%%%%%%%%%%%%%%%%%%%%%%%%%%%%%%%%%%%%%%%%%%%%%%%%
% Experimental Setup
%%%%%%%%%%%%%%%%%%%%%%%%%%%%%%%%%%%%%%%%%%%%%%%%%%%%%%%%%%%%%%%%%%%%%%%%%%%%%%%%
% ----------------------
\section{Experimental setup}\label{sec:exp-defs}
%
% [JRH] TODO: What more can I say to provide more info on experimental
% evaluation, per ijcai comment? Also address ijcai#10 comment
% about unclear robot capabilities, implementation.
We use the ARGoS simulator~\citep{Pinciroli2012}~\footnote{Our open-source code
  is available at https://github.com/swarm-robotics/fordyca,
  https://github.com/swarm-robotics/titerra} with a dynamical physics model of
the marXbot robot in a 3D space for maximum fidelity (robots are still
restricted to motion in the XY plane).

Robots run the FSM shown in~\cref{fig:fsm}, use short range proximity sensors
for collision avoidance, and ground sensors to check if they are currently on
top of a block. Blocks are abstracted as black squares on the arena floor.  We
ran four sets of experiments characterized by different swarm density, and
evaluate our model across the scenarios shown in~\cref{fig:foraging-dists},
relaxing some core assumptions made when deriving our model as we proceed.

% [JRH] Address ijcai comment about lack of obstacles
We ran our experiments with $\TheSwarmSize=1\dots6,006$ robots in environments with SS
and DS source block distributions, and with $\TheSwarmSize=1\dots11,837$ robots for RN
and PL block distributions. Our scenarios are obstacle-free in the sense that
they do not contain external obstacles; however, robots act as obstacles to each
other. We use $\theta=\frac{\pi}{36}$ for CRW and compute $\tau_{av}$ from the
details of our robot control algorithm $\kappa$ (details omitted).
$\ScenarioCACharacterization{\TheScenario}$ is computed post-hoc for each
experiment. For all experiments we performed 32 runs of $T=200,000$ seconds,
with a timestep of $t=0.2$ seconds, i.e., the robot control algorithm runs at 5
Hz.

%%%%%%%%%%%%%%%%%%%%%%%%%%%%%%%%%%%%%%%%%%%%%%%%%%%%%%%%%%%%%%%%%%%%%%%%%%%%%%%%
% Results
%%%%%%%%%%%%%%%%%%%%%%%%%%%%%%%%%%%%%%%%%%%%%%%%%%%%%%%%%%%%%%%%%%%%%%%%%%%%%%%%
%
%%% Local Variables:
%%% mode: latex
%%% TeX-master: "2021-ar-main"
%%% End:
% ----------------------
\section{Results}\label{sec:results}
For all the scenarios, we omit predictions for $\SwarmNSearching$ since it can
be computed from the conservation of robots via $\SwarmNSearching =
\TheSwarmSize - \SwarmNAvoiding - \SwarmNHoming$. We present our results with a
non-scaled X-axis, with one data point per experiment, and a logarithmic Y-axis
for larger $\TheSwarmSize$, to improve readability.

% ----------------------
\subsection{Constant $\SwarmDensity$, large $\TheSwarmSize$}\label{ssec:results-large-const-rho}
These are the ``ideal'' conditions under which to evaluate our model, because
the linearity and mean-field assumptions we made when deriving our model
hold.

    \begin{figure}[t]
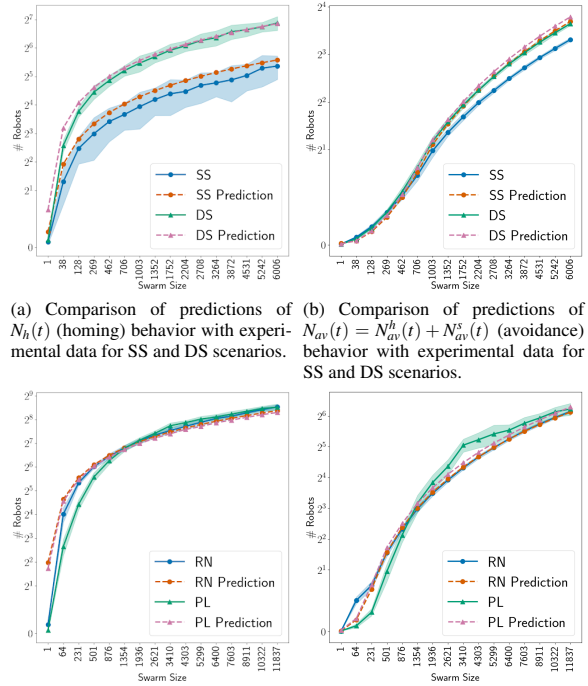

  \begin{minipage}[t]{.49\linewidth}
    {\includegraphics[width=\linewidth]{\figroot{graphs_hetero_const}/large_const_ss.png}}
    \footnotesize{(a) Comparison of predictions of $\SwarmNHoming$ (homing) behavior with
      experimental data for SS and DS scenarios.}
  \end{minipage}
  \hfill
  \begin{minipage}[t]{.49\linewidth}
    {\includegraphics[width=\linewidth]{\figroot{large_hetero}/SS.16x8x2+DS.16x8x2-sc-graphs/sc-N_av-d0.CRW.png}}
    \footnotesize{(b) Comparison of predictions of
      $\SwarmNAvoiding=\SwarmNAvoidingWhileHoming+\SwarmNAvoidingWhileSearching$
      (avoidance) behavior with experimental data for SS and DS scenarios.\rule[-1ex]{0ex}{1ex}}
  \end{minipage}
  \begin{minipage}[t]{.49\linewidth}
    {\includegraphics[width=\linewidth]{\figroot{graphs_hetero_const}/large_const_rn.png}}
    \footnotesize{(c) Comparison of predictions of $\SwarmNHoming$ (homing) behavior with
      experimental data for RN and PL scenarios.}
  \end{minipage}
  \hfill
  \begin{minipage}[t]{.49\linewidth}
    {\includegraphics[width=\linewidth]{\figroot{large_hetero}/RN.8x8x2+PL.8x8x2-sc-graphs/sc-N_av-d0.CRW.png}}
    \footnotesize{(d) Comparison of predictions of
      $\SwarmNAvoiding=\SwarmNAvoidingWhileHoming+\SwarmNAvoidingWhileSearching$
      (avoidance) behavior with experimental data for RN and PL scenarios.\rule[-1ex]{0ex}{1ex}}
  \end{minipage}
  \caption{\label{fig:results-large-const-rho}Predictions of swarm
      behavior at large scales with constant swarm density $\SwarmDensity=0.1$
      for single source (SS), dual source (DS), random (RN), and power law (PL)
      scenarios.}
\end{figure}
In~\cref{fig:results-large-const-rho}, we see strong agreement between the
predictions of our model and experimental results for all scenarios, providing
compelling evidence that our model is capturing the underlying dynamics of
inter-robot interference and searching accurately and that our mean-field and
linearity assumptions are accurate. PL scenarios are the least favorable of all
foraging environments, since they are asymmetrical and do not contain easily
exploitable block clusters. Our model struggles to predict
$\SwarmNHoming,~\SwarmNAvoiding$ within the 95\% confidence interval for PL
scenarios, but does track the overall trend reasonably well, showing that our
underlying diffusion model and assumptions about linearity of $\alpha_r$ are
generally accurate even in this difficult case. The divergence that does exist
between predictions and experiments for PL scenarios suggests that
\cref{eqn:acq-pdf-single} is moderately inaccurate; this is further supported by
slight differences between experimental data and predictions for $\SwarmNHoming$
in the RN and SS scenarios

% Do you mean "scenarios' environments" or "scenario environments"? 

environments.
In~\cref{fig:results-large-const-rho}b, when our predictions of
$\SwarmNAvoiding$ are inaccurate for some DS scenarios (in terms of confidence
intervals), the inaccuracy may not be of practical concern as the predictions
only differ from experimental results by $< 1$ robot/$6,000$.

% ----------------------
\subsection{Constant $\SwarmDensity$, small $\TheSwarmSize$}\label{ssec:results-small-const-rho}
We relax the mean-field assumption, and evaluate our model's ability to capture
finite-size effects when $\TheSwarm$ cannot be reasonably approximated using the
mean-field model (with $\TheSwarmSize\le{50}$).

    \begin{figure}[t]
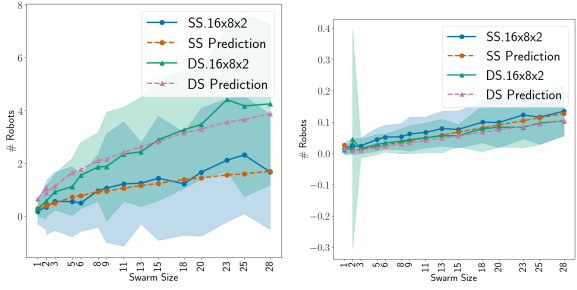
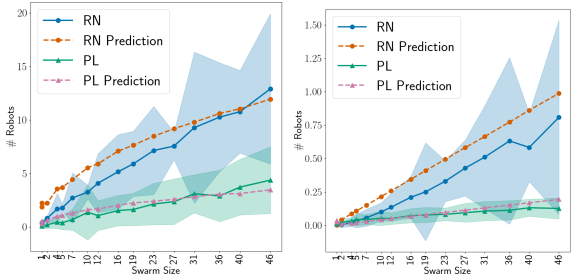

  \begin{minipage}[t]{.49\linewidth}
    {\includegraphics[width=\linewidth]{\figroot{graphs_hetero_const}/small_const_updated_ss_ds.png}}
    \footnotesize{(a) Comparison of predictions of $\SwarmNHoming$ (homing) behavior with
      experimental data for SS and DS scenarios.}
  \end{minipage}
  \hfill
  \begin{minipage}[t]{.49\linewidth}
    % \subfloat[\label{fig:perf-rn-pl}
    {\includegraphics[width=\linewidth]{\figroot{small_hetero}/SS.16x8x2+DS.16x8x2-sc-graphs/sc-N_av-d0.CRW.png}}
    \footnotesize{(b) Comparison of predictions of
      $\SwarmNAvoiding=\SwarmNAvoidingWhileHoming+\SwarmNAvoidingWhileSearching$
      (avoidance) behavior with experimental data for SS and DS scenarios.\rule[-1ex]{0ex}{1ex}}
  \end{minipage}
  %\hfill
  \begin{minipage}[t]{.49\linewidth}
    {\includegraphics[width=\linewidth]{\figroot{graphs_hetero_const}/small_const_updated_rn_pl.png}}
    \footnotesize{(c) Comparison of predictions of $\SwarmNHoming$ (homing)
      behavior with experimental data for RN and PL scenarios.}
  \end{minipage}
  %\hfill
  \begin{minipage}[t]{.49\linewidth}
    {\includegraphics[width=\linewidth]{\figroot{small_hetero}/RN.8x8x2+PL.8x8x2-sc-graphs/sc-N_av-d0.CRW.png}}
    \footnotesize{(d) Comparison of predictions of
      $\SwarmNAvoiding=\SwarmNAvoidingWhileHoming+\SwarmNAvoidingWhileSearching$
      (avoidance) with experimental data for RN and PL scenarios.\rule[-1ex]{0ex}{1ex}}
  \end{minipage}
  \caption{\label{fig:results-small-const-rho}\footnotesize{Predictions of swarm
      behavior at small scales with constant swarm density $\SwarmDensity=0.1$
      for single source (SS), dual source (DS), random (RN), and power law (PL)
      scenarios.}}
\end{figure}
In~\cref{fig:results-small-const-rho}, we see strong agreement between
experimental results and the predictions of our model for all tested scenarios
with small $\TheSwarmSize$. Even when this crucial assumption is relaxed, our
model accurately captures finite-size effects, suggesting that the mean-field
assumption may be valid at very small $\TheSwarmSize$, depending on the control
algorithm used. The greatest discrepancy between predictions and results is
shown in~\cref{fig:results-small-const-rho}c for RN scenarios, which are
inherently the most non-linear because of the placement of the nest relative to
the objects to be gathered.  In comparison with results from the previous
section, stronger agreement between predictions and results with constant
$\SwarmDensity$ is seen when $\TheSwarmSize$ is small, likely due to the
inherent stochasticity of swarms which comes into play more strongly at larger
$\TheSwarmSize$.

% ----------------------
\subsection{Variable $\SwarmDensity$, small $\TheSwarmSize$}\label{ssec:results-small-var-rho}
We relax the linearity \emph{and} mean-field assumptions, and evaluate our
model's ability to capture finite-size effects when $\TheSwarm$ exhibits
non-linear behaviors with $\SwarmDensity=0.01-0.1$ and $\TheSwarmSize\le{50}$.

    \begin{figure}[t]
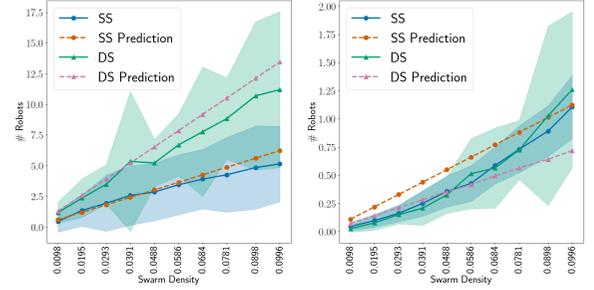
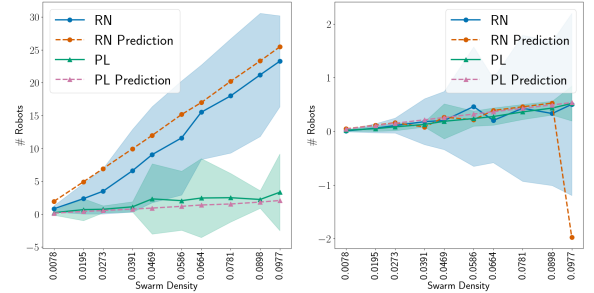

  \begin{minipage}[t]{.49\linewidth}
    {\includegraphics[width=\linewidth]{\figroot{graphs_hetero_var}/small_var_updated_ss_ds.png}}
    \footnotesize{(a) Comparison of predictions of $\SwarmNHoming$ (homing) behavior with
      experimental data for SS and DS scenarios.}
  \end{minipage}
  \hfill
  \begin{minipage}[t]{.49\linewidth}
    {\includegraphics[width=\linewidth]{\figroot{small_hetero}/SS.32x16x2+DS.32x16x2-sc-graphs/sc-N_av-d0.CRW.png}}
    \footnotesize{(b) Comparison of predictions of
      $\SwarmNAvoiding=\SwarmNAvoidingWhileHoming+\SwarmNAvoidingWhileSearching$
      (avoidance) behavior with experimental data for SS and DS scenarios.\rule[-1ex]{0ex}{1ex}}
  \end{minipage}
  \hfill
  \begin{minipage}[t]{.49\linewidth}
    {\includegraphics[width=\linewidth]{\figroot{graphs_hetero_var}/small_var_updated_rn_pl.png}}
    \footnotesize{(c) Comparison of predictions of $\SwarmNHoming$ (homing) behavior with
      experimental data for RN and PL scenarios.}
  \end{minipage}
  \hfill
  \begin{minipage}[t]{.49\linewidth}
    % \subfloat[\label{fig:perf-rn-pl}
    {\includegraphics[width=\linewidth]{\figroot{small_hetero}/RN.16x16x2+PL.16x16x2-sc-graphs/sc-N_av-d0.CRW.png}}
    \footnotesize{(d) $\SwarmNAvoiding=\SwarmNAvoidingWhileHoming+\SwarmNAvoidingWhileSearching$
      (avoidance) with experimental data for RN and PL scenarios.\rule[-1ex]{0ex}{1ex}}
  \end{minipage}
  \caption{\label{fig:results-small-var-rho}\footnotesize{Predictions of swarm
      behavior at small scales with variable swarm density
      $\SwarmDensity=0.01-0.1$ for single source (SS), dual source (DS), random
      (RN), and power law (PL) scenarios. Swarm size $\TheSwarmSize$ varies
      $5\dots51$ for SS and DS scenarios, and $2\dots25$ for RN and PL
      scenarios.}}
\end{figure}
In~\cref{fig:results-small-var-rho} we see that our model generally performs
well even when we violate its low $\SwarmDensity$ assumption. We see strong
agreement between the model and experimental data for all scenarios for
$\SwarmDensity=0.01-0.05$ for both $\SwarmNHoming$ and
$\SwarmNAvoiding$. In~\cref{fig:results-small-var-rho}d we see a striking
divergence between our model and results at $\SwarmDensity=0.09$, which may be a
numerical anomaly or the point at which our ODE solver begins to struggle with
the non-linear behavioral inputs.

% ----------------------
\subsection{Variable $\SwarmDensity$, large $\TheSwarmSize$}\label{ssec:results-large-var-rho}
We restore the mean-field assumption to evaluate our model's ability to capture
collective dynamics across scales on the order of 1,000s of robots when
$\TheSwarm$ exhibits non-linear behaviors. We believe that $\SwarmDensity=0.1$
is an appropriate upper limit for real-world swarms of practical utility, as it
corresponds to 1 robot/$10m^2$ which is an extremely high density in the context
of feasible real-world applications.

    \begin{figure}[t]
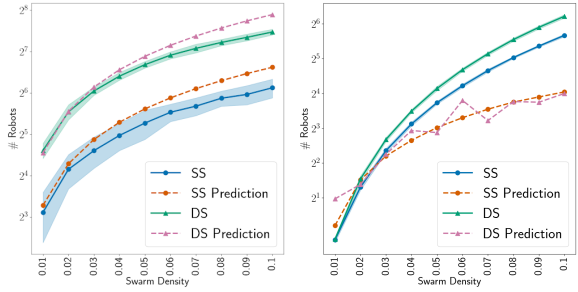
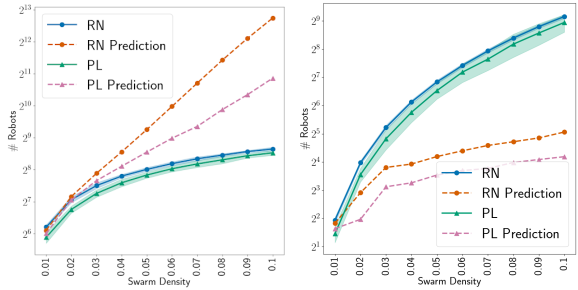

  \begin{minipage}[t]{.49\linewidth}
    {\includegraphics[width=\linewidth]{\figroot{graphs_hetero_var}/large_var_ss.png}}
    \footnotesize{(a) Comparison of predictions of $\SwarmNHoming$ (homing) behavior with
      experimental data for SS and DS scenarios.}
  \end{minipage}
  \hfill
  \begin{minipage}[t]{.49\linewidth}
    {\includegraphics[width=\linewidth]{\figroot{large_hetero}/SS.256x128x2+DS.256x128x2-sc-graphs/sc-N_av-d0.CRW.png}}
    \footnotesize{(b) Comparison of predictions of
      $\SwarmNAvoiding=\SwarmNAvoidingWhileHoming+\SwarmNAvoidingWhileSearching$
      (avoidance) behavior with experimental for SS and DS scenarios.\rule[-1ex]{0ex}{1ex}}
  \end{minipage}
  \hfill
  \begin{minipage}[t]{.49\linewidth}
    {\includegraphics[width=\linewidth]{\figroot{graphs_hetero_var}/large_var_rn.png}}
    \footnotesize{(c) Comparison of predictions of $\SwarmNHoming$ (homing) behavior with
      experimental data for RN and PL scenarios.}
  \end{minipage}
  \hfill
  \begin{minipage}[t]{.49\linewidth}
    {\includegraphics[width=\linewidth]{\figroot{large_hetero}/RN.256x256x2+PL.256x256x2-sc-graphs/sc-N_av-d0.CRW.png}}
    \footnotesize{(d) Comparison of predictions of
      $\SwarmNAvoiding=\SwarmNAvoidingWhileHoming+\SwarmNAvoidingWhileSearching$
      (avoidance) behavior with experimental data for RN and PL scenarios.\rule[-1ex]{0ex}{1ex}}
  \end{minipage}
  \caption{\label{fig:results-large-var-rho}Predictions of swarm behavior at
    large scales with variable swarm density
    $\SwarmDensity=0.01-0.1$. $\SwarmNHoming$ and $\SwarmNAvoiding$ shown for
    single source (SS), dual source (DS), random (RN), and power law (PL)
    scenarios, random (RN) and power law (PL) scenarios. Swarm size
    $\TheSwarmSize$ varies $327\dots3276$ for SS and DS scenarios, and
    $655\dots6553$ for RN and PL scenarios.}
\end{figure}

In~\cref{fig:results-large-var-rho}, we see that our model generally does not
model swarm behavior at large scales well. In~\cref{fig:results-large-var-rho}
we see divergence between model predictions and results for
$\SwarmNAvoiding,~\SwarmNHoming$ for all scenarios beyond $\SwarmDensity=0.02$,
and beyond $\SwarmDensity=0.01$ for RN, PL scenarios, and our model fails to
provide meaningful predictions, due to the highly non-linear swarm behaviors
present, which is as we expected. For RN and PL scenarios, we see a strong
correlation between the $\SwarmDensity$ at which our model's predictions for
both $\SwarmNHoming$ and $\SwarmNAvoiding$ break down, which happens at
$\approx\SwarmDensity=0.01$.

%%%%%%%%%%%%%%%%%%%%%%%%%%%%%%%%%%%%%%%%%%%%%%%%%%%%%%%%%%%%%%%%%%%%%%%%%%%%%%%%
% Discussion
%%%%%%%%%%%%%%%%%%%%%%%%%%%%%%%%%%%%%%%%%%%%%%%%%%%%%%%%%%%%%%%%%%%%%%%%%%%%%%%%
%
\section{Discussion}\label{sec:discussion}

% [JRH] Address ijcai#5 comment about not providing empirical evidence
% that our model is better than previous work. Address ijcai#10 comment
% that our model probably doesn't capture dynamics at both large and small
% scales (also in the results section).
Our results % in~\cref{sec:results}
advance the state of the art in ODE modeling
for foraging swarms~\citep{Lerman2001,Lerman2003a} by demonstrating
accurate predictions of swarm behaviors across scales and block
distributions when the assumptions made in~\cref{ssec:method-preliminaries}
are valid. To the best of our knowledge, our model is the first presented in the
literature capable of capturing the collective dynamics of swarms when the
mean-field approximation is valid, i.e., large $\TheSwarmSize$, but also when the
mean-field approximation is not valid and finite-size effects are present. i.e., small
$\TheSwarmSize$. Building on our results, accurate predictions % from first principles
of swarm
performance and SR system properties, such as scalability and emergent
self-organization using~\citep{Harwell2020a}, are possible.

\begin{figure}[t]
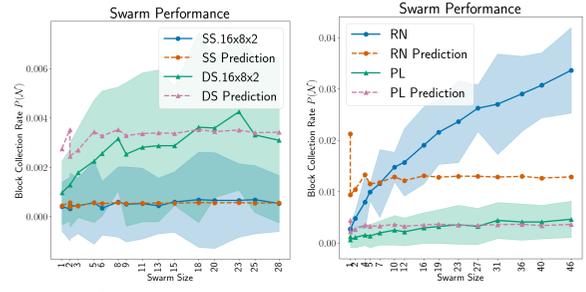

\begin{minipage}[t]{0.49\linewidth}
  {\includegraphics[width=\linewidth]{\figroot{small_hetero}/SS.16x8x2+DS.16x8x2-sc-graphs/sc-PM-ss-raw-d0.CRW.png}}
  \footnotesize{(a) SS and DS:  $\TheSwarmSize={1}\ldots{50}$.}
\end{minipage}
\hfill
\begin{minipage}[t]{0.49\linewidth}
%\centering
  {\includegraphics[width=\linewidth]{\figroot{small_hetero}/RN.8x8x2+PL.8x8x2-sc-graphs/sc-PM-ss-raw-d0.CRW.png}}
  \footnotesize{(b) RN and PL:  $\TheSwarmSize={1}\ldots{50}$.\rule[-1ex]{0ex}{1ex}}
\end{minipage}
\caption{\label{fig:perf-predict} Predictions of swarm performance from first
  principles for different scenarios}
\end{figure}

As an example, we can %directly
predict swarm performance with reasonable accuracy
using our model, as shown in~\cref{fig:perf-predict}. We define a performance
measure $P$ as the rate of block collection by the swarm $\TheSwarm$. We then
predict it via $P=\alpha_{b}/\mu_{h}$, where $\mu_{h}$ is the rate of robots
leaving the homing queue $\QueueHoming$ (analogous to $\QueueCA$). Under our
assumption of $\QueueHoming$ as a M/M/1 queue, we set
$\mu_{h}=1$. In~\cref{fig:perf-predict} we see strong agreement between our
predictions and experimental results for SS and PL scenarios across all swarm
sizes, and reasonable agreement for DS scenarios. Our predictions are less
accurate for RN scenarios, likely due to our approximation of $\QueueCA$ as an
M/M/1 queue not being correct; i.e., our model does not account for multiple
robots dropping an object in the nest on a single $t$.
%

% Maria -- can you be more specific on which cases in Fig 4 the model struggles?
% [JRH] Edited to be more clear on this point.
When we consider the swarm behavior with constant $\SwarmDensity$ at large
$\TheSwarmSize$ (\cref{fig:results-large-const-rho}), we see our model struggle
to predict behavior for \emph{all} portions of the curve for all scenarios. We
see that while our model predicts behavior at $\TheSwarmSize \ge 100$ well, it
struggles to predict behavior at $\TheSwarmSize<100$ for SS, DS, and RN
scenarios, underscoring the difficulty in creating a single linear model which
can capture both mean-field and finite-size behaviors simultaneously. PL
scenarios are the most challenging to model, and we see that while our model
follows the general trend of behavior well, its predictions are usually outside
of the confidence interval range. However, our model \emph{is} accurate to
within $\approx16$ robots/$11,837$, which may be sufficient in many cases.

Our linearity assumption, strictly speaking, is clearly not valid in all
contexts across scales or across scenarios at similar
$\TheSwarmSize$. Nevertheless, it is valid in \emph{many} contexts (e.g.,
\cref{fig:results-small-const-rho,fig:results-small-var-rho}, parts
of~\cref{fig:results-large-const-rho}), and thus our model is of practical
utility in designing SR systems to tackle problems across scales.  Furthermore,
our findings suggest that $\TheSwarmSize$ \emph{always} has a non-linear effect
on behavior, even at extremely low densities, which is a function of the
scenario itself. For more favorable scenarios, this non-linearity may be
negligible even at large $\TheSwarmSize$, but for less favorable scenarios, such
as RN and PL, it is not, and non-linear modeling must be applied for accurate
prediction across scales.

When we relax our linearity assumption and increase $\SwarmDensity$ to introduce
non-linear behaviors in the swarm via inter-robot interference, our model still
produces comprehensively accurate predictions of behavior at small
$\TheSwarmSize$ for all scenarios (\cref{fig:results-small-var-rho}).  While our
model will undoubtedly break down at some $\SwarmDensity > 0.1$, our results
demonstrate that the threshold is above any realistic scenario, because
densities $ > 0.1$ are infeasible in general for real-world problems.

Capturing non-linear behaviors using linear models is much more successful if
$\TheSwarmSize$ is small; this aligns well with our intuition.  If we imagine
that the amount of time each robot spends avoiding collision with other robots
at each timestep ($t_{ca}$) is roughly of the form $t_{ca} =
\TheSwarmSize^{\SwarmDensity}$ then it is clear that $t_{ca}$ will increase more
quickly with $\TheSwarmSize$, even if $\SwarmDensity$ is low.

Based on this discussion, we can confidently characterize the limits of linear
modeling, at least for ODE based approaches, and say that it is \emph{possible}
to capture non-linear behaviors of swarms using linear models across
scales. Furthermore, for $\TheSwarmSize\le{50}$, doing so is also
\emph{practical}, as our model still provided predictions accurate within 95\%
confidence in all scenarios. However, scenarios exhibiting comparatively
\emph{more} non-linear behaviors via $\SwarmNAvoiding$, such as RN, have large
95\% confidence intervals, reflecting the greater variability in behavior as a
result of non-linearity. As a result, our model's predictions of the swarm
average behavior may be less useful in such scenarios, since the model provides
predictions of the average case. We note that while the results of this work
\emph{suggest} that linear modeling of non-linear swarm behaviors is possible
using other modeling approaches, they do not necessarily generalize to other
modeling paradigms.

Our model still relies on one parameter:
$\ScenarioCACharacterization{\TheScenario}$. While this is a free parameter in
the strictest sense, it has a strong semantic context rather than just being a
fitting parameter to ODE terms to improve prediction accuracy. This is a
substantial improvement over previous work which used five context-free fitting
parameters~\citep{Lerman2002}.  Thus, while our model is not (yet) strictly
fundamental, the semantic context of $\ScenarioCACharacterization{\TheScenario}$
strongly suggests that it \emph{can} be determined analytically, and
first-principle modeling of general foraging behavior is possible.

Finally, it is important to emphasize the presented results were obtained in
simulation, commensurate with other mathematical approaches to modeling SR
systems~\citep{GuerreroBonilla2020b,Tarapore2017}. Even if our model does not
produce equivalently accurate results to those shown
in~\cref{fig:results-small-const-rho,fig:results-small-var-rho} with systems of
real robots, we believe that it will still provide a strong starting point for
modeling systems of real robots, thereby accelerating the deployment of swarm
solutions to real-world problems, for two reasons. First, the robot FSM used in
this work (\cref{fig:fsm}) is simple: it has minimal sensor usage, and does not
use memory. This simplicity will make real-world systems of such robots
especially amenable to mathematical modeling.
%because fewer simplifying assumptions will need to be made to model their
%behavior.
Second, our model has
% Maria -- can it be more precise on the size?  In Fig 5 and 8 the number of
% robots is small, while in fig 4 is very large. When the x-axis is density (
% fig 6 and 7), how many robots are used? The paper says 6,000 or 8,000 but it
% is not clear for what cases.
% [JRH] Updated to make this clearer.
been shown to be accurate with swarms of non-trivial size (e.g., up to 46 robots
in~\cref{fig:results-small-const-rho}, up to 51 robots
in~\cref{fig:results-small-var-rho}, parts of~\cref{fig:results-large-const-rho}
with up to 11,837 robots). In such swarms, non-deterministic transient behaviors
will reliably arise even in simulation, e.g., floating point representation
errors; similar errors will arise even in small swarms of real robots.
%% Although our model might not cope with them out of the box, it has the
%% capability to do so, as demonstrated in this work.

\section{Conclusions and future work}\label{sec:conclusion}
We have investigated how to characterize the limits of linear modeling of swarm
behaviors. We have developed a robust ODE model for foraging swarms, and shown
it is accurate in a wide variety of scenarios in swarms which exhibit linear
behaviors, and are large enough to be approximated using mean-fields,
demonstrating its utility as an effective modeling tool. Furthermore, in swarms
for which either assumption is not true, our model in many cases is still
accurate. Future work will derive the block acquisition density function using
random walks theory and refine our model with experiments using real robots.

\noindent{\bf Acknowledgments:}
We gratefully acknowledge the MnDRIVE initiative, the Minnesota Robotics
Institute, the UofM Informatics Institute, and the Minnesota Supercomputing
Institute for their support of this research.

%
%%%%%%%%%%%%%%%%%%%%%%%%%%%%%%%%%%%%%%%%%%%%%%%%%%%%%%%%%%%%%%%%%%%%%%%%%%%%%%%%
% Bibliography
%%%%%%%%%%%%%%%%%%%%%%%%%%%%%%%%%%%%%%%%%%%%%%%%%%%%%%%%%%%%%%%%%%%%%%%%%%%%%%%%
% \bibliographystyle{spmpsci}
%\bibliographystyle{plainnat}
%\bibliographystyle{sn-basic}
\bibliography{2022-ode,2022-phd-thesis}

\end{document}